\documentclass[conference]{IEEEtran}
\usepackage{cite}
\usepackage{amsmath,amssymb,amsfonts}
\usepackage{algorithmic}
\usepackage{graphicx}
\usepackage{textcomp}
\usepackage{xcolor}
\usepackage{caption}
\usepackage{subfigure}
\usepackage{blindtext}
\usepackage{gensymb}
\usepackage{url}
\usepackage[absolute]{textpos}
\usepackage[pagebackref=true,breaklinks=true,letterpaper=true,colorlinks,bookmarks=false]{hyperref}
\def\BibTeX{{\rm B\kern-.05em{\sc i\kern-.025em b}\kern-.08em
    T\kern-.1667em\lower.7ex\hbox{E}\kern-.125emX}}

\begin{document}
\begin{textblock}{20}(1,0.2)
\noindent\small This article has been accepted for publication in 2022 International Conference on Innovation and Intelligence for Informatics, Computing, and \\ Technologies (3ICT). Citation information - DOI: 10.1109/3ICT56508.2022.9990696, 3ICT-IEEE (\href{https://ieeexplore.ieee.org/document/9990696}{https://ieeexplore.ieee.org/document/9990696})
\end{textblock}
%


\title{Short-Term Aggregated Residential Load Forecasting using BiLSTM and CNN-BiLSTM \\
\thanks{*Note: Only the personal use of this preprint is permitted - solely for the purpose of research, but republication/redistribution requires IEEE permission. Check \protect\url{http://www.ieee.org/publications_standards/publications/rights/index.html} for more information. Please cite this paper from this link: \href{https://ieeexplore.ieee.org/document/9990696}{https://ieeexplore.ieee.org/document/9990696}.}
}

\author{\IEEEauthorblockN{\small Bharat Bohara, \textit{Student Member, IEEE}}
\IEEEauthorblockA{\small ECE Department\\
\small\textit{University of Houston}\\
\small Houston, United States \\
\small bbohara@uh.edu}
\and
\IEEEauthorblockN{\small Raymond I. Fernandez}
\IEEEauthorblockA{\small ECE Department\\
\small \textit{University of Houston}\\
\small Houston, United States \\
\small rifernan@uh.edu}
\and
\IEEEauthorblockN{\small Vysali Gollapudi}
\IEEEauthorblockA{\small ECE Department\\
\small \textit{University of Houston}\\
\small Houston, United States \\
\small vgollap2@uh.edu}
\and
\IEEEauthorblockN{\small Xingpeng Li, \textit{Senior Member, IEEE}}
\IEEEauthorblockA{\small ECE Department\\
\small \textit{University of Houston}\\
\small Houston, United States \\
\small xli82@uh.edu}
}

\IEEEoverridecommandlockouts
\IEEEpubid{\makebox[\columnwidth]{978-1-6654-5193-2/22/\$31.00~\copyright2022 IEEE \hfill} \hspace{\columnsep}\makebox[\columnwidth]{ }}

\maketitle

\begin{abstract}
Higher penetration of renewable and smart home technologies at the residential level challenges grid stability as utility-customer interactions add complexity to power system operations. In response, short-term residential load forecasting has become an increasing area of focus. However, forecasting at the residential level is challenging due to the higher uncertainties involved. Recently deep neural networks have been leveraged to address this issue. This paper investigates the capabilities of a bidirectional long short-term memory (BiLSTM) and a convolutional neural network-based BiLSTM (CNN-BiLSTM) to provide a day ahead ($24$ hr.) forecasting at an hourly resolution while minimizing the root mean squared error (RMSE) between the actual and predicted load demand. Using a publicly available \href{https://zenodo.org/record/5642902#.Y-Wsp3bMJPY}{dataset} consisting of $38$ homes, the BiLSTM and CNN-BiLSTM models are trained to forecast the aggregated active power demand for each hour within a $24$ hr. span, given the previous $24$ hr. load data. The BiLSTM model achieved the lowest RMSE of 1.4842 for the overall daily forecast. In addition, standard LSTM and CNN-LSTM models are trained and compared with the BiLSTM architecture. The RMSE of BiLSTM is $5.60\%$, $2.85\%$ and $2.60\%$ lower than LSTM, CNN-LSTM and CNN-BiLSTM models respectively. The source code of this work is available at \href{https://github.com/Varat7v2/STLF-BiLSTM-CNNBiLSTM.git}{https://github.com/Varat7v2/STLF-BiLSTM-CNNBiLSTM.git}.
\end{abstract}


\section{Introduction}
\IEEEPARstart{A}{} load forecasting is a technique used by the power/energy utilities to estimate the future load profiles so as to balance the power demand-supply chain. A short-term load forecasting (STLF) ranges from estimating the load demand for the next half an hour up to the next two weeks. An accurate load forecasting techniques help the utilities to plan or schedule the power generations optimally. It has a greater economic impact on the utilities as it plays a critical role in optimal load scheduling, economic load dispatch, superlative load flow, and contingency analysis, and intelligent planning, operations, and the maintenance of the power systems \cite{Jacob2020}. Indeed, the economy of operations and control of power systems are sensitive to positive and negative forecasting errors. Under-estimates have a negative impact on demand response and power system operations, whereas overload conditions challenging to manage. Over-estimates impacts power system operation and hence the system’s efficiency \cite{Chen2010}. Even $1\%$ reduction in the average STLF error can potentially save a typical utility company millions of dollars \cite{Alfares2010}. Thus, STLF has emerged as an essential area of research for efficient and reliable power system operation.

At the residential level, higher penetration of renewables, integration of smart home technologies, and the electrification of both vehicles and heating/cooling systems present a challenge to grid stability as utility-customer interactions add complexity to power system operations. In response, short-term residential load forecasting has become an increasing area of focus \cite{Ahmad2019}. However, forecasting at this level, even for aggregated loads, is challenging because of the diverse electricity consumption behaviors and patterns. Thus it adds up a rapid fluctuation and variability in the data, making the prediction job more challenging. Residential household load profiles vary significantly depending on the house size, occupancy, presence of solar panels, electric vehicle usage, and other socio-demographic factors \cite{Haben2019}. STLF for smaller residential communities is a complex process because of the non-smooth and nonlinear temporal load profile behaviors. Thus, accurate methods for STLF at the residential level are needed and will provide valuable information for the effective generation, and distribution of the electricity \cite{Amjady2010}.

Machine learning techniques are widely becoming popular in Machine learning techniques are widely popular in predicting the aggregated residential electric load. Zhang \textit{et al.} \cite{Zhang2015} have applied a combination of state vector machine (SVM) based clustering and decision tree techniques to smart meter data to build a short-term load forecasting framework. The model can accurately predict the power usage pattern with real-time operation constraints. Stephen \textit{et al.} \cite{Stephen2017} have applied a Gaussian-Markov chain sampling technique to smart meter data to predict residential power usage. Hence, machine learning-based models achieved equivalent accuracy but were observed to be computationally expensive compared to a traditional autoregressive integrated moving average (ARIMA) model.

Deep learning has become an active area of research in short-term aggregated residential load forecasting. Specifically, long short-term memory (LSTM) architectures are becoming an increasing area of focus. One of the first attempts to use an LSTM was conducted by Marino \textit{et al.} \cite{Marino2016}. Standard LSTM and LSTM-based sequence-to-sequence (S2S) architectures were employed to predict the aggregated electricity consumption of residential customers. The study demonstrated that the S2S LSTM architecture could accurately forecast loads with both one-minute and one-hour temporal resolution. This result marked a notable improvement compared to previous studies with mostly shallow artificial neural networks. Similar studies have addressed residential short-term load forecasting as well. Kong \textit{et al.} \cite{Kong2019} have tested an LSTM neural network on residential smart meter data and compared the performance to a comprehensive set of benchmarks in the field of load forecasting. Hence, from the above studies, it is observed that the LSTM approach outperforms other state-of-the-art forecasting algorithms in forecasting short-term loads.

Other studies have addressed the more challenging task of individual residence load forecasting. Wang \textit{et al.} \cite{Wang2021} have studied a short-term residential load forecasting using an LSTM model that takes a two-dimensional feature of load and weather as input, and a mean absolute percentage error (MAPE) score of $48.46\%$ is achieved. This marked a $9.8\%$ accuracy improvement compared to a MAPE score of $53.77\%$ achieved using the one-dimensional load input. In addition to the multiple feature input data, researchers have also experimented with more complex LSTM architectures. Alhussein \textit{et al.} \cite{Alhussein2020} has proposed a deep learning framework consisting of combined convolutional neural network (CNN) and LSTM networks. The hybrid CNN-LSTM model is tested on individual household electrical load data, and a better result with an average MAPE of $40.38\%$ is achieved compared to other forecasting techniques. In this study, MAPE score of $8.18\%$ and $44.6\%$ is achieved for aggregated and individual household load predictions, respectively. Hence, because of the possibility of better model accuracy, an aggregated household load is chosen over the individual household for further analytical comprehension.

There is very limited work done on short-term load forecasting using BiLSTM and CNN-BiLSTM. Wu \textit{et al.} \cite{Wu2021} has proposed an attention-based CNN-LSTM-BiLSTM model to forecast one hour ahead load profiles of regional integrated energy systems with historical load, cooling load, temperature, and gas consumption for the past 5 days as the input features. The authors have stated that the proposed method outperformed all other models such as CNN-BiLSTM, CNN-LSTM, BiLSTM, LSTM, random forest regression (RFR), and support vector machine regression (SVR). Similarly, Miao \textit{et al.} \cite{Miao2021} has proposed a short-term load forecasting with CNN-BiLSTM for highly accurate time series prediction with Bayesian Optimization (BO) to tune the model hyperparameters and Attention Mechanism (AM) to focus on the important part of the BiLSTM layers. The proposed model takes historical load profiles, time slots and meteorological data to predict load for the next 24 hours. The attention mechanism-based learning is not included in our work, however, it can bookmarked for the future work.

This paper proposes a bidirectional LSTM (BiLSTM) and convolutional BiLSTM (CNN-BiLSTM) to predict the aggregated household power consumption in a small residential community. The main idea is to forecast a day ahead load profile, given the previous day's load at an hourly resolution. The contribution of this work is to push the potential of the regular and convolutional BiLSTM further to model the short-term forecasting at the lower time resolution without any significant loss in the model performance.

\section{Background}
\subsection{LSTM Architecture}
Recurrent Neural Networks (RNNs) are a branch of deep neural networks that are designed to process sequential data \cite{Goodfellow-et-al-2016}. One of the main problems faced in RNN is \textit{gradient vanishing} issue. With more depth of neural layers, the gradients assigned to the weight matrix go on diminishing and finally have no effect on the output. The other problem observed in conventional RNNs is \textit{gradient exploding} - again, with the increasing number of hidden layers, the gradients assigned to the weight matrices grow exponentially, thus making output incomprehensible. Both the gradient vanishing and the gradient exploding make RNNs hard to train.

 \begin{figure}[h]
    \centering
	\includegraphics[width=0.5\textwidth]{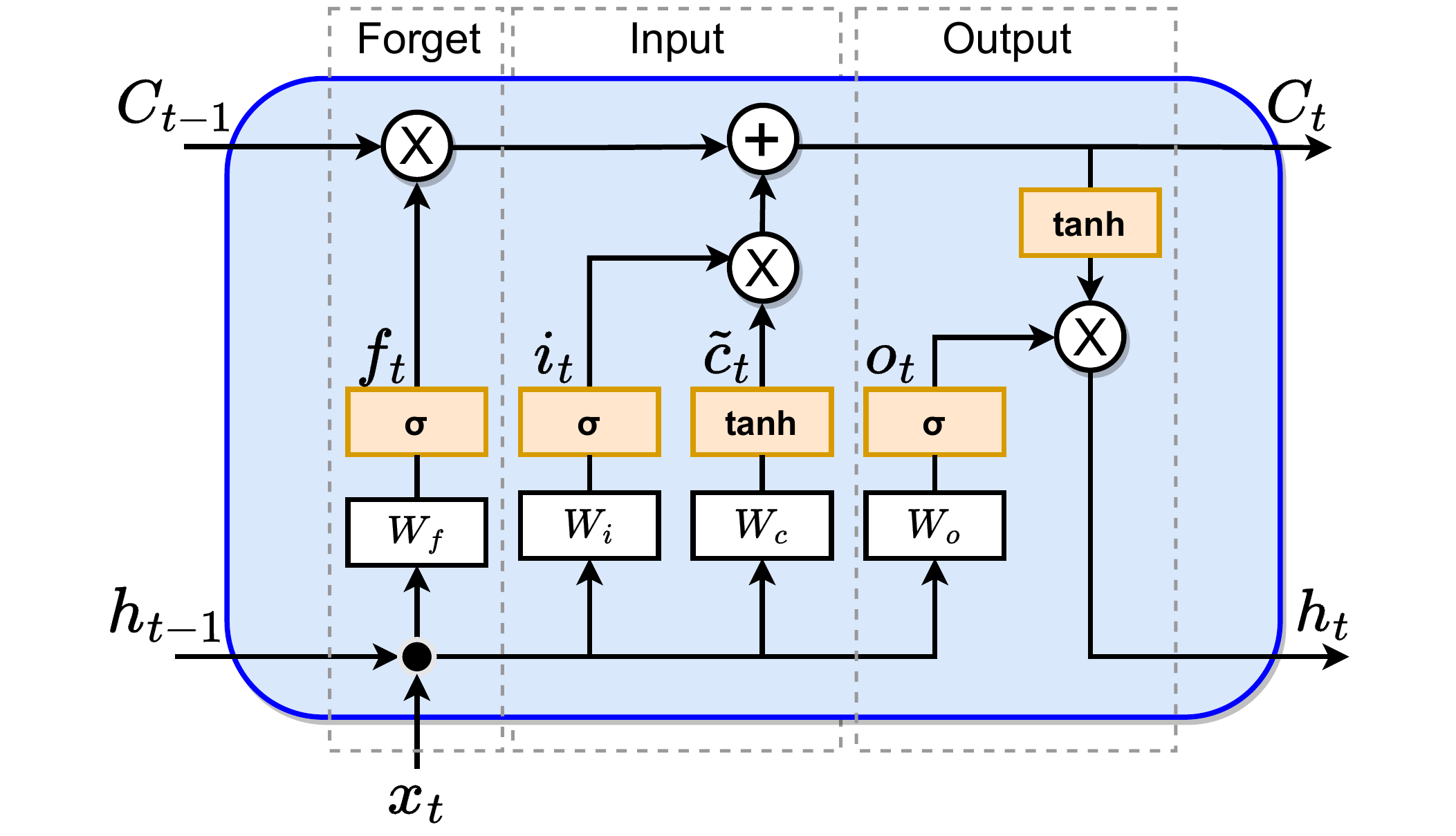}
	\captionsetup{font=small}
	\caption{LSTM cell with gating controls}
	\label{fig:lstm_cell}
\end{figure}

LSTM - a gated RNN, overcomes the difficulty faced by traditional RNNS in learning the long-term dependencies of the sequential input data \cite{Siami-Namini2019}. It can leverage the long-term previous states sequential information with the involvement of gates - forget gate ($f_t$), input gate ($i_t$), and output gate ($o_t$). The LSTM cell architecture with the gating controls is shown in fig. \ref{fig:lstm_cell}, where all the gates $f_t, i_t, o_t \;\epsilon\; \{0,1\}$: $0$ means that the gate is closed and no information passes through it, and $1$ means the gate is open to allow the information flow. The forget gate ($f_t \;\epsilon\; \{0,1\}$) is constituted of a sigmoid function that determines whether to retain or discard the information from the cell memory, i.e., $0$ - indicates forgetting and $1$ - indicates remembering the cell memory information. The input or update gate is composed of two layers: 1) a sigmoid function ($i_t \;\epsilon\; \{0,1\}$) - that determines whether to update the cell input or not, and 2) a $\tanh$ function ($\tilde{c} \;\epsilon\; \{-1,1\}$) - that creates a vector of new candidate values to be added to the memory cell. At the end, the combination of both the layers and the forget layer updates the LSTM memory cell as shown in equation \ref{eqn:lstm-cell-output}. The output gate is also composed of a sigmoid, and a $\tanh$ function: $\sigma \;\epsilon\; \{0,1\}$ determines if LSTM-cell information has contributed to the overall cell output, and hyperbolic tangent function maps the final cell output between $-1$ and $1$. In other words, the output gate controls the output of the information and determines if the current hidden state will be passed to the next sequence network. The mathematical framework of a LSTM-cell is depicted in equations \ref{eqn:lstm-cell}, \ref{eqn:lstm-cell-vectorized}, \ref{eqn:lstm-cell-generalized}, and \ref{eqn:lstm-cell-output} \cite{Cai2019, Siami-Namini2019, Yu2020}.

\begin{equation}
    \begin{gathered}
        i_t = \sigma (w_{ix} x_t + w_{ih} h_{t-1} + b_i) \\
        f_t = \sigma (w_{fx} x_t + w_{fh} h_{t-1} + b_f) \\
        \tilde{c}_t = \tanh (w_{cx} x_t + w_{ch} h_{t-1} + b_c) \\
        o_t = \sigma (w_{ox} x_t + w_{oh} h_{t-1} + b_o) \\
    \end{gathered}
    \label{eqn:lstm-cell}
\end{equation}

\noindent where, $i_t$ is a input gate, $f_t$ is a forget gate, $\tilde{c_t}$ is a candidate hidden state, and $o_t$ is an output gate of a LSTM cell. Eqn. \ref{eqn:lstm-cell} can be rearranged in a vectorized form as shown in eqn. \ref{eqn:lstm-cell-vectorized}.

\begin{equation}
    \begin{gathered}
        \begin{bmatrix}
        i_t \\
        f_t \\
        o_t
    \end{bmatrix}
    = \sigma \left(
    \begin{bmatrix}
        w_{ix} & w_{ih} \\
        w_{fx} & w_{fh} \\
        w_{ox} & w_{oh}
    \end{bmatrix}
    \cdot
    \begin{bmatrix}
        x_t \\
        h_{t-1}
    \end{bmatrix}
    +
    \begin{bmatrix}
        b_i \\
        b_f \\
        b_o
    \end{bmatrix}
    \right)\\
    \begin{bmatrix}
        \tilde{c_t}
    \end{bmatrix}
    = \tanh \left(
    \begin{bmatrix}
        w_{cx} & w_{ch}
    \end{bmatrix}
    \cdot
    \begin{bmatrix}
        x_t \\
        h_{t-1}
    \end{bmatrix}
    +
    \begin{bmatrix}
        b_c
    \end{bmatrix}
    \right)
    \end{gathered}
    \label{eqn:lstm-cell-vectorized}
\end{equation}

\noindent The vectorized eqn. \ref{eqn:lstm-cell-vectorized} can be condensed to the eqn. \ref{eqn:lstm-cell-generalized}.

\begin{equation}
    \begin{gathered}
        \mathbf{y_1} = \sigma(\mathbf{w_1 x_1}+\mathbf{b_1}) \\
        \mathbf{y_2} = \tanh(\mathbf{w_2 x_2}+\mathbf{b_2})
    \end{gathered}
    \label{eqn:lstm-cell-generalized}
\end{equation}

\noindent where, \\

\noindent $\mathbf{y_1} =
    \begin{bmatrix}
        i_t & f_t & o_t
    \end{bmatrix}^T$,
$\mathbf{w_1} = 
    \begin{bmatrix}
        w_{ix} & w_{fx} & w_{ox} \\
        w_{ih} & w_{fh} & w_{oh}
    \end{bmatrix}^T$, \\
$\mathbf{x_1} =
    \begin{bmatrix}
        x_t & h_{t-1}
    \end{bmatrix}^T$, \;\;
$\mathbf{b_1} =
    \begin{bmatrix}
        b_i & b_f & b_o
    \end{bmatrix}^T$ \\
$\mathbf{y_2} =
    \begin{bmatrix}
        \tilde{c_t}
    \end{bmatrix}$, \qquad\qquad\;
$\mathbf{w_2} = 
    \begin{bmatrix}
        w_{cx} & w_{ch}
    \end{bmatrix}^T$, \\
$\mathbf{x_2} =
    \begin{bmatrix}
        x_t & h_{t-1}
    \end{bmatrix}^T$, \;\;\,
$\mathbf{b_2} =
    \begin{bmatrix}
        b_c
    \end{bmatrix}$ \\


\noindent The final cell output at each time step will be as shown in eqn. \ref{eqn:lstm-cell-output}.
\begin{equation}
    \begin{gathered}
        c_t = f_t \ast c_{t-1} + i_t \ast \tilde{c} \\
        h_t = o_t \ast \tanh (c_t)
    \end{gathered}
    \label{eqn:lstm-cell-output}
\end{equation}
\noindent where, $c_t$ is cell states and $h_t$ is a vector of cell output values.

\subsection{BiLSTM Architecture}
\begin{figure}[h]\centering
\includegraphics[width=0.5\textwidth]{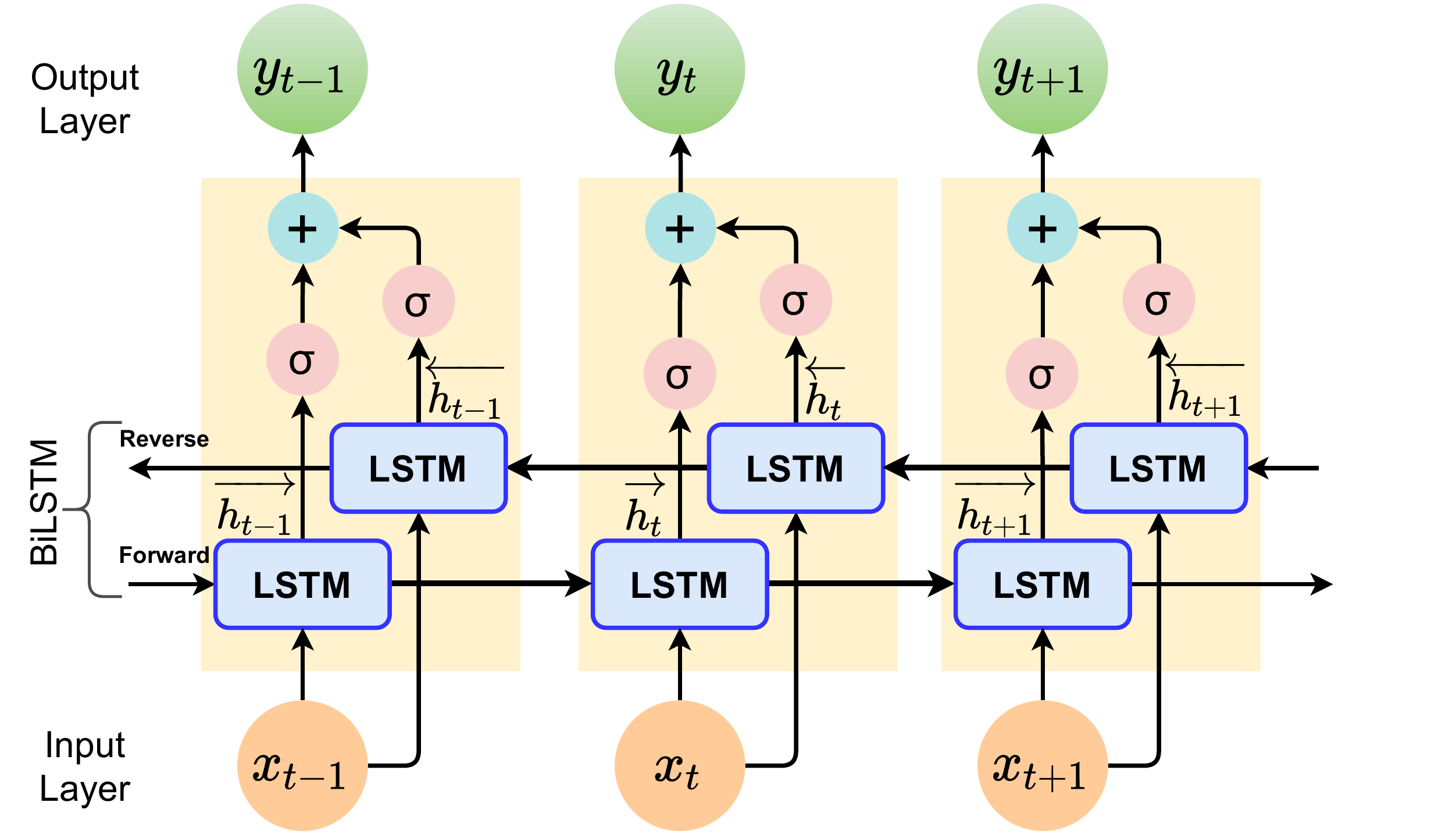}
\captionsetup{font=small}
\caption{Architecture of the BiLSTM network}
\label{fig:bilstm-cell}
\end{figure}

The BiLSTM neural network architecture used in this study is illustrated in Fig. \ref{fig:bilstm-cell}. To overcome the shortcoming of a single LSTM cell that can capture only the previous state information but not the future state information, a Bi-LSTM model is used in this paper. It is built upon two independent LSTM networks: a forward LSTM cell that passes information from the back to the front direction and a reverse LSTM cell that passes information from the front to back direction. This allows the network to leverage the learning from both the forward and reverse direction of the input sequence at every time step. Through this flexibility, the network can look at both the previous and future context information to predict the next step sequence.

A BiLSTM cell is fed with a input sequence of $x=(x_1, x_2, \dots, x_n)$, where $n$ represent the length of the sequence. In reference to a BiLSTM cell architecture shown in fig. \ref{fig:bilstm-cell}, $\overrightarrow{h_t}=(\overrightarrow{h_1}, \overrightarrow{h_2}, \dots, \overrightarrow{h_n})$ is forward hidden sequence, $\overleftarrow{h_t}=(\overleftarrow{h_1}, \overleftarrow{h_2}, \dots, \overleftarrow{h_n})$ is reverse hidden sequence, and $y_t=(y_1, y_2, \dots, y_n)$ is the output sequence. The final encoded output vector is the combined effect of both forward and reverse information flow i.e., $y_t=f(\overrightarrow{h_t}, \overleftarrow{h_t})$ \cite{Cai2019}. The mathematical framework of the BiLSTM neural networks architecture is shown in eqn. \ref{eqn:bilstm-cell}.

\begin{equation}
    \begin{gathered}
        \overrightarrow{h_t} = \sigma (w_{\overrightarrow{h}x} x_t + w_{\overrightarrow{h}\overrightarrow{h}} h_t + b_{\overrightarrow{h}}) \\
        \overleftarrow{h_t} = \sigma (w_{\overleftarrow{h}x} x_t + w_{\overleftarrow{h}\overleftarrow{h}} h_t + b_{\overleftarrow{h}}) \\
        y_t = w_{y\overrightarrow{h}} \overrightarrow{h_t} + w_{y\overleftarrow{h}} \overleftarrow{h_t} + b_y
    \end{gathered}
    \label{eqn:bilstm-cell}
\end{equation}

\subsection{CNN-BiLSTM Architecture}
A convolution neural network (CNN) is cascaded on top of the BiLSTM networks to extract the prior feature knowledge of the inputs before fetching them into the BiLSTM layers. Although CNN layers add additional computational cost, the CNN-BiLSTM model is not observed to perform equivalent to the BiLSTM. Its poor performance might be because of losing some information due to the max-pooling layers.

\begin{figure}[ht]
\centering
\captionsetup{justification=centering}
\includegraphics[width = 0.49\textwidth]{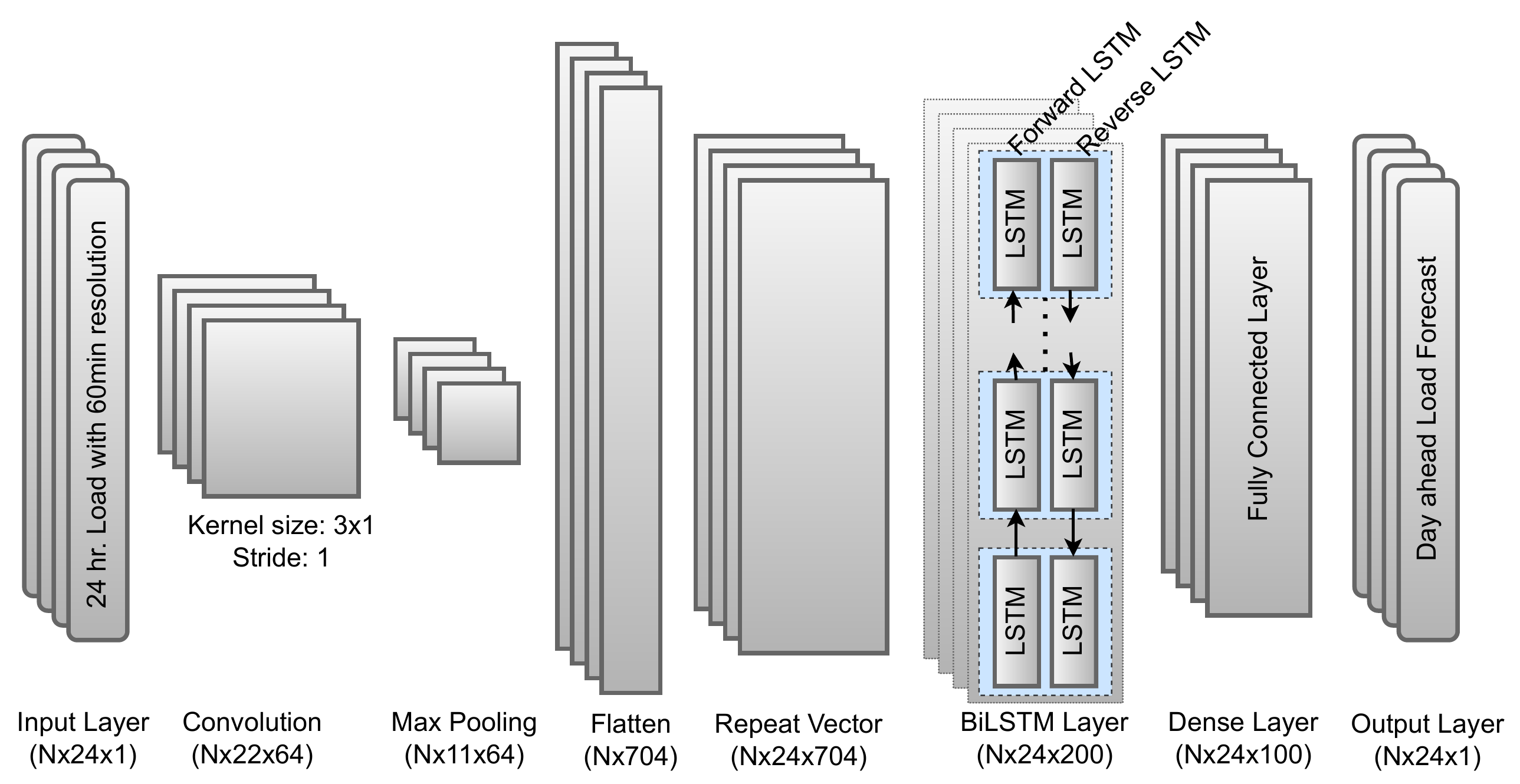}
\captionsetup{font=small}
\caption[]{Architecture of the CNN-BiLSTM network}
\label{fig:cnn-bilstm}
\end{figure}

As shown in fig. \ref{fig:cnn-bilstm}, first, the entire dataset is sliced into per day ($24$ hr.) load profiles, and each slice is considered a single input to the $1D$ CNN layer. $64$ filters with a kernel size of $3\text{x}1$ and a default stride of $1$ are used to extract the feature vectors from the input load profile data. Max-pooling is applied on the convoluted feature maps. It reduces the dimension but also might have lost some vital information. It is the reason behind its lower accuracy compared to the BiLSTM networks. In order to perform sequential load demand forecasting, BiLSTM networks are used on the flattened feature vectors. Finally, the load profile for the next day is outputted by the fully-connected layers with the $ReLU$ activation function. The only difference in CNN-LSTM is that it uses regular forward-directed LSTM layers instead of the bidirectional layers for sequential time-series predictions.

\subsection{Naive Forecasting Technique}
The main idea of the naive forecasting technique is that it assumes the previous day's load profile as the forecast for the next day's load demand. This technique applies only to data with a periodic repeating and recurring sequence pattern. Since the hourly load profile data has a strong repetitive seasonal component, as observed in fig. \ref{fig:seasonal_decomposition}, this technique is used in this study for comparison purposes. On the other hand, the naive forecasting technique is usually used in any recurring sequential time-series data because of its simplicity and inexpensive computational cost.
\begin{equation}
    D(t+1) = D(t) \quad \forall \;\, t=\{1,2,3, \dots \, n\}
    \label{eqn:naive-forecast}
\end{equation}
\noindent where, $D$ is an hourly load profile for a particular day, $t$ is a previous day, $t+1$ is the next day and $n$ is the total length of the sequence in days.

\section{Methodology}
\subsection{Dataset Preparation}
\begin{figure}[h]
    \centering
    \captionsetup{justification=centering}
	\includegraphics[width=0.4\textwidth]{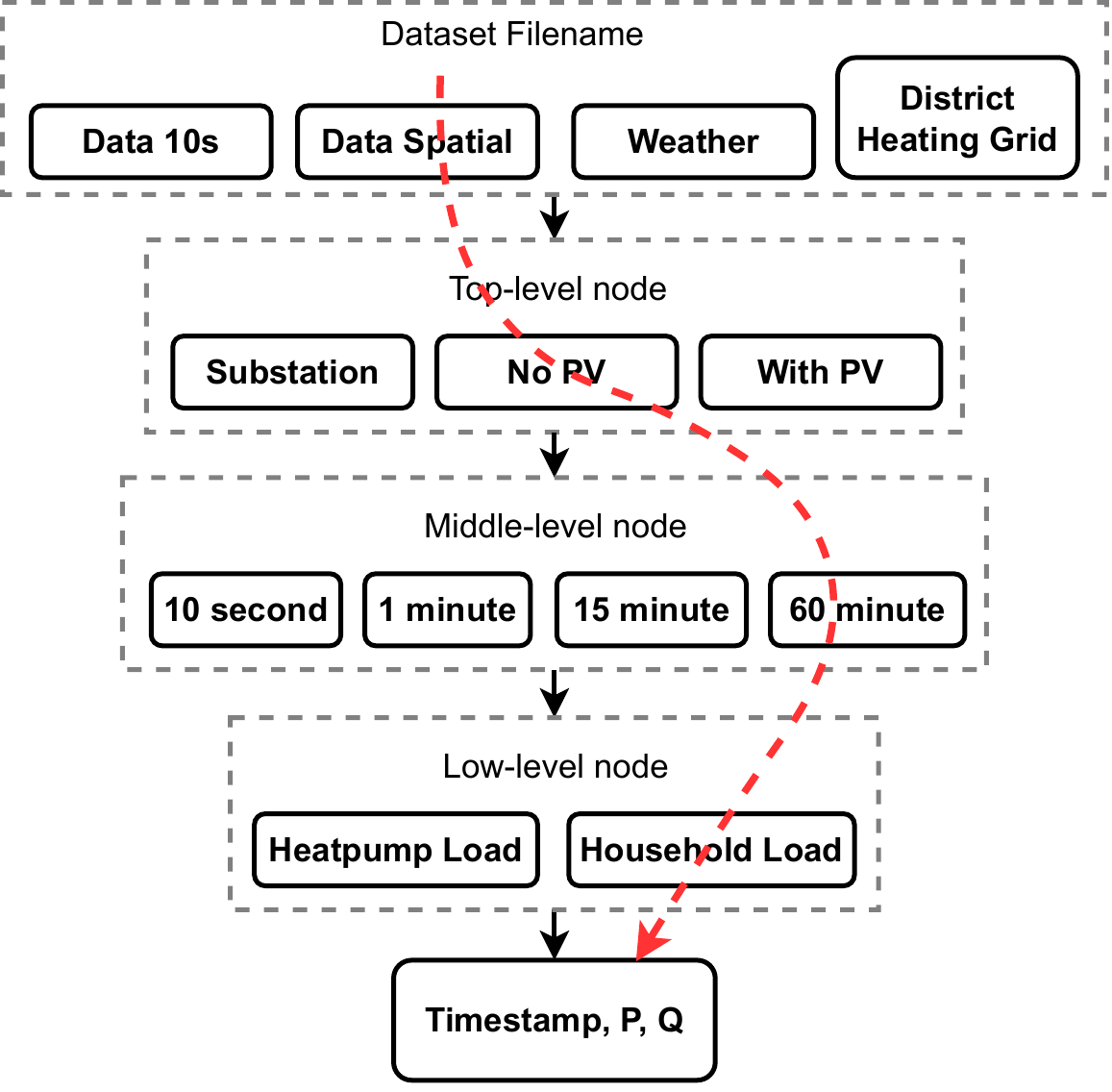}
	\captionsetup{font=small}
	\caption{Hierarchy of dataset published in \cite{Schlemminger2022}, where red-dashed arrow indicates the data-subset used in this study}
	\label{fig:data_hierarchy}
\end{figure}

An aggregated residential electrical load profiles measured in $38$ single-family houses in Northern Germany \cite{Schlemminger2022} is used in this study. The dataset consists of active and reactive power per household from $10$ seconds to $60$ minute temporal resolution dated from May $2018$ to December of $2020$. The data is open-sourced and consists of seven hierarchical data format version 5 (HDF5) files for each year - that consists of different hierarchies as shown in fig. \ref{fig:data_hierarchy}. The dataset used in this study is a subset of the larger HDF5 dataset published in \cite{Schlemminger2022}; the subset is generated with the red-dashed arrowed direction as shown in fig. \ref{fig:data_hierarchy} - with filename \textit{data\_spatial}, top-level node \textit{NO\_PV}, middle-level node \textit{60min}, and low-level node \textit{HOUSEHOLD}. The generated subset dataset consists of a timestamp and active and reactive power. For load forecasting, only active power in $kW$ is used. The dataset is divided into training, testing, and validation sets based on the date, as shown in the table \ref{tab:data_split}. A Min-max normalization eqn. \ref{eqn:normalization} is applied to the training, validation, and test dataset before feeding into the models.

\begin{equation}
    \text{X}_{norm} = \frac{\text{X}-\text{X}_{min}}{\text{X}_{max}-\text{X}_{min}}
    \label{eqn:normalization}
\end{equation}

\noindent where $\text{X}$ is a regular and $\text{X}_{norm}$ is the normalized data.

\begin{table}[h]
\centering
\caption{Data split into training, validation, and test set}
\label{tab:data_split}
\begin{tabular}{|c|c|c|}
\hline
Category & Load Profile From & Load Profile To \\ \hline
Training set & 5/31/2018 & 12/31/2019 \\ \hline
Validation set & 1/1/2020 & 6/30/2020 \\ \hline
Test set & 7/1/2020 & 12/31/2020 \\ \hline
\end{tabular}
\end{table}

\begin{figure}[h]
\centering
\captionsetup{justification=centering}
\includegraphics[width=0.49\textwidth]{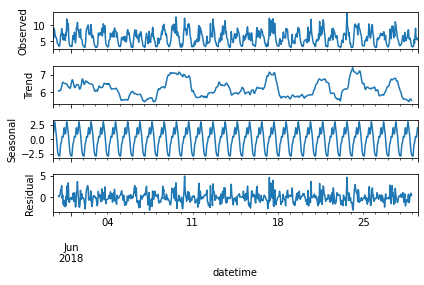}
\captionsetup{font=small}
\caption[]{Time-series decomposition of the data}
\label{fig:seasonal_decomposition}
\end{figure}

An additive time series decomposition is applied to the dataset in order to split the time series data into several components such as seasonality, trend, and noise. Seasonality gives the periodic signal in the time series data; trend gives the hidden pattern if present in the data, and noise gives the remaining random signals. Each of these components will be helpful in data preparation, model selection, and model finetuning. A representative example of an arbitrary $30$ day period of data is additively decomposed, as depicted in fig. \ref{fig:seasonal_decomposition}, at an hourly resolution. The data has a strong seasonality and significant noise, but no distinct trend exists. A complex trend and a noise component make it difficult to forecast the power demand with typical statistical techniques. However, the strong daily seasonality pattern suggests that a naive forecast made by using a day before load to predict the day ahead load profile can provide a good forecast baseline measurement.

\subsection{Objective and Metrics}
The main objective of the load forecasting technique is to minimize the error between the actual and predicted daily load forecasts at an hourly resolution. Root mean squared error (RMSE) metric is used to evaluate the performance of the trained model.
\begin{equation}
\begin{aligned}
J(\theta) = \min_{w,b} \quad & \frac{1}{n} \sqrt{\sum_{i=1}^{n} \left( y_i - \hat{y_i}\right)^2}
\end{aligned}
\end{equation}

\noindent where $n \in \{1,2,3, \dots, 24\}$ is the total number of data points for $24$ hrs. with an hourly resolution, $y_i$ is the actual value at $i^{th}$ hour, $\hat{y_i}$ is the predicted value at $i^{th}$ hour, the cost function may be terminated via an early stopping technique to prevent the model from overfitting, and parameters $\theta \in \{w, b\}$ are weights and biases of the neural networks that are learned during the model training with the observation data as shown in table \ref{tab:data_split}.

\subsection{Proposed Method}
A dataset consisting of the aggregated household load is first divided into training, validation, and test sets. Training and validation sets are used to train the BiLSTM and CNN-BiLSTM networks. The test set is held back to evaluate the model's performance. Training and validation sets are normalized and then fed into the model. The model initially takes in the first 24 hrs. of data and attempts to predict the following 24 hrs. load profile. In the second training iteration, the window is shifted ahead by 1 hour. The window shifting continues until the dataset is completely exhausted. Simultaneously, the validation set evaluates the model performance and monitors the issues such as overfitting or underfitting. The model accuracy can be further improved by increasing the training set and varying the batch size and the total number of BiLSTM cells and layers. Adam optimizer adaptively optimizes the learning rate based on the error value and the rate of change of the error. Once the model is trained, a walk-forward test strategy is employed to evaluate the model's performance. During the model evaluation, the model initially takes in the first 24 hrs. of test data and uses the trained model to predict the following 24 hrs. load profile. During the second test iteration, again, the window is shifted ahead by 24 hrs. time frame throughout the available test data points.

\subsection{Hyperparameter Tuning}
The hyperparameters such as batch size, number of BiLSTM cells, and the network layer size are tuned by using a grid search method. The batch size and the number of total cells in the BiLSTM layer are varied, and the corresponding RMSE values are reported in table \ref{tab:rmse_grid_search}. The parameters - batch size of $384$ and LSTM cells of $200$, resulted in the lowest average RSME value of $1.4842$ with minimum and maximum values ranging between [$1.4800,1.5020$]. However, with multiple folds grid search method, these tuned parameters resulted in the lowest RMSE value multiple times. Hence these parameters are finally considered for model training.

\begin{table}[h]
\centering
\captionsetup{font=small}
\caption{RMSE Values for Grid Search Parameters}
\label{tab:rmse_grid_search}
\begin{tabular}{|c|c|c|c|}
\hline
\textbf{Batch size} & \textbf{150 cells} & \textbf{200 cells} & \textbf{400 cells} \\ \hline
\textbf{96} & 1.5053 & 1.5023 & 1.5240 \\ \hline
\textbf{192} & 1.5518 & 1.5279 & 1.5039 \\ \hline
\textbf{384} & 1.4981 & \textbf{1.4842} & 1.5171 \\ \hline
\end{tabular}
\end{table}

\section{Results and Discussion}
\subsection{Naive Forecasting Technique}
\begin{figure}[h]
\centering
\captionsetup{justification=centering}
\subfigure[]{\includegraphics[width = 0.24\textwidth]{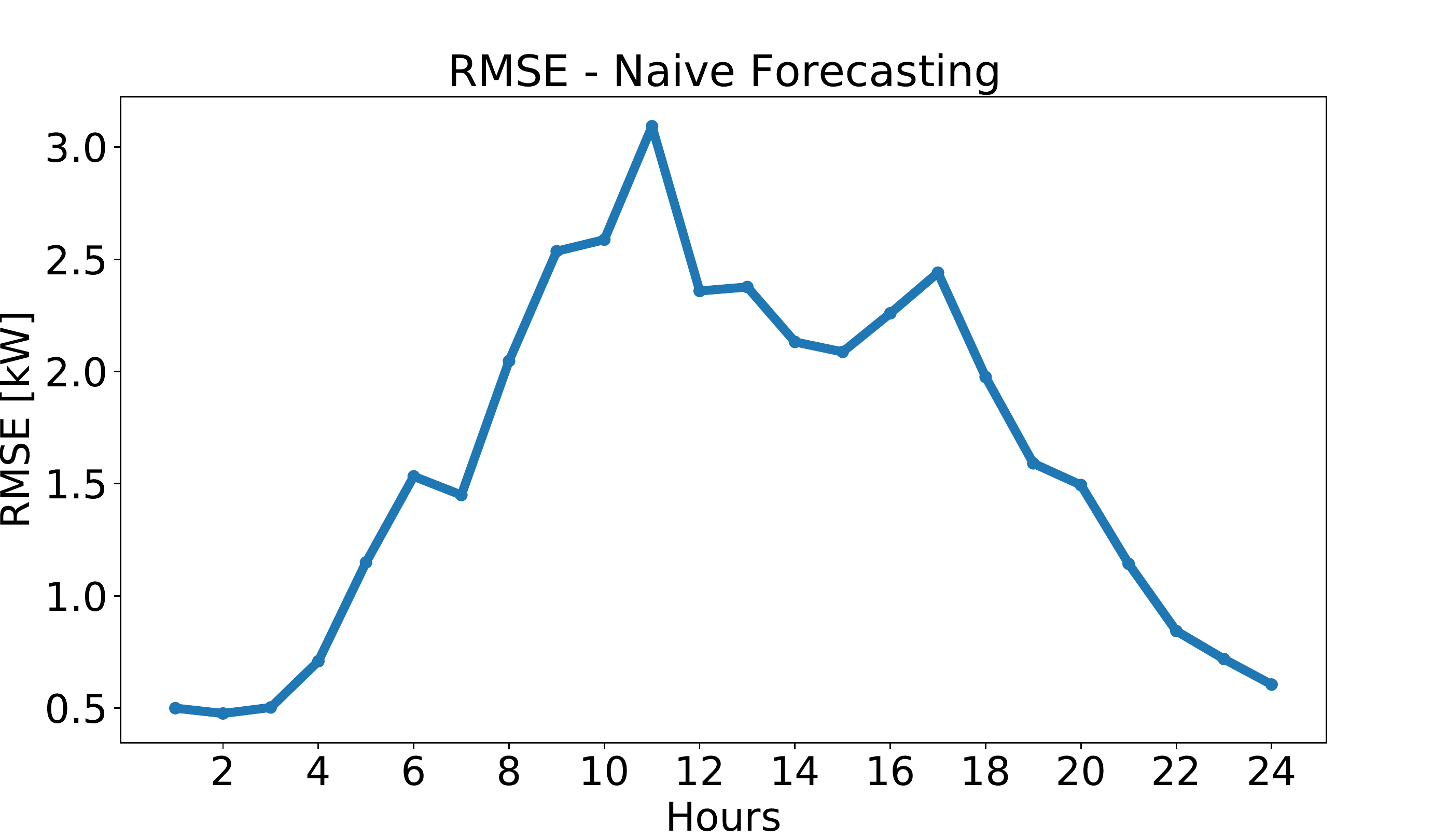}}
\subfigure[]{\includegraphics[width = 0.24\textwidth]{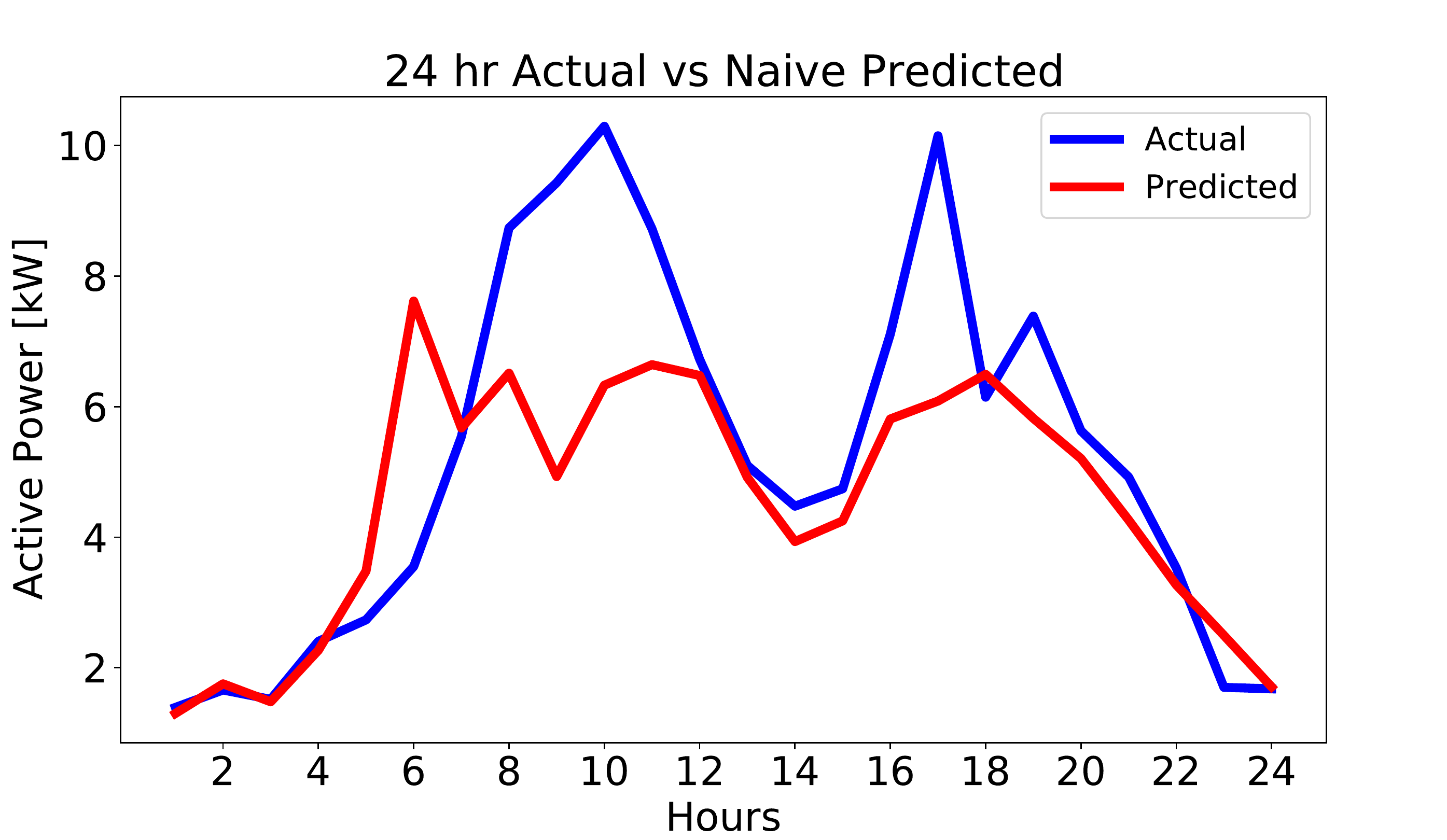}}
\captionsetup{font=small}
\caption[]{(a) RMSE and (b) predicted load profile using naive forecasting technique}
\label{fig:naive-plots}
\end{figure}

Since the naive forecasting technique is generally adopted in real-life forecasting problems, it is considered the baseline in this study. It is, by default, human nature to assume that similar events will get repeated in the subsequent days. The hourly RMSE error between the actual and predicted load demand is calculated as shown in fig. \ref{fig:naive-plots}(a), and the resulting prediction output is shown in fig. \ref{fig:naive-plots}(b). Lower error values indicate better prediction is observed in the early morning and late-night hours compared to the mid-day. An average daily RSME value for the naive forecasting technique is observed to be $1.7282 kW$.

\subsection{BiLSTM Forecast}
\begin{figure}[h]
\centering
\captionsetup{justification=centering}
\subfigure[]{\includegraphics[width = 0.24\textwidth]{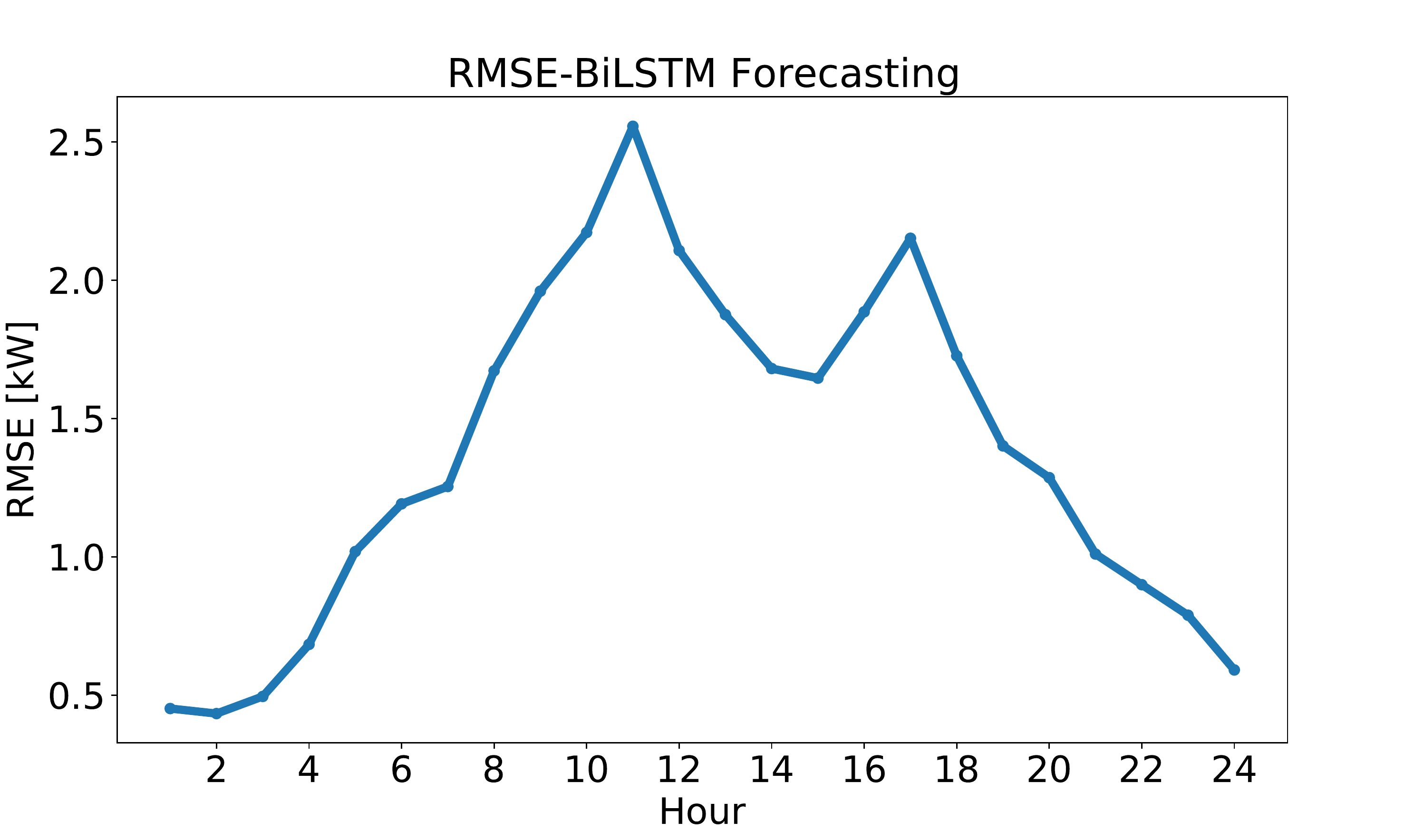}}
\subfigure[]{\includegraphics[width = 0.24\textwidth]{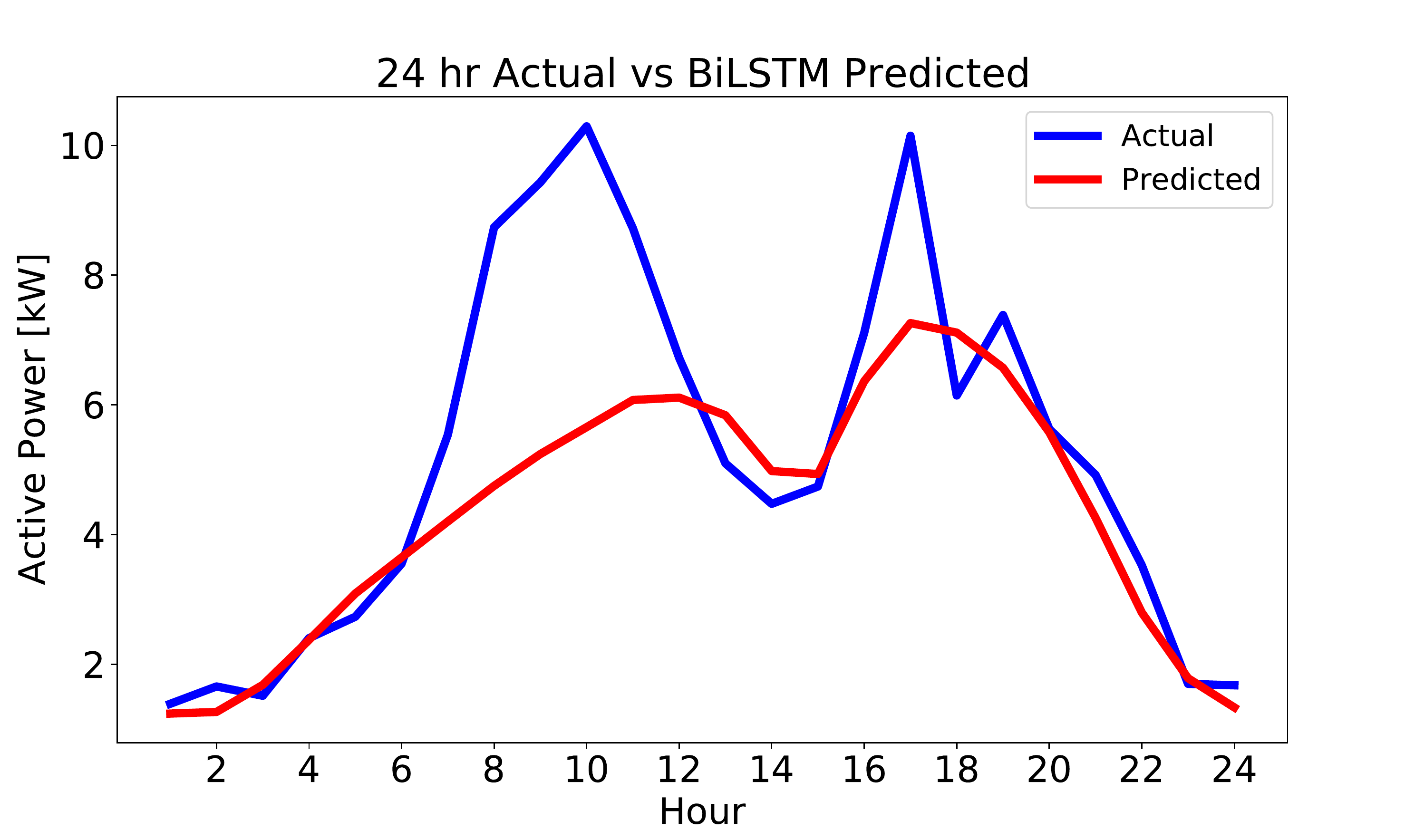}}
\captionsetup{font=small}
\caption[]{(a) RMSE between actual and predicted load, (b) hourly forecasted load profile using BiLSTM netwroks}
\label{fig:bilstm-plots}
\end{figure}

Fig. \ref{fig:bilstm-plots}(a) shows the hourly RSME plot between the actual and predicted load demand. Like the naive forecasting technique, the lower RMSE values indicate better predictions are observed in the early morning and late-night hours. An average daily RMSE score for the BiLSTM forecast method is observed to be $1.4842 kW$, which shows a $13.53\%$ decrease in the average error compared to the naive forecast.

\begin{figure}[h]
\centering
\captionsetup{justification=centering}
\subfigure[]{\includegraphics[width = 0.2412\textwidth]{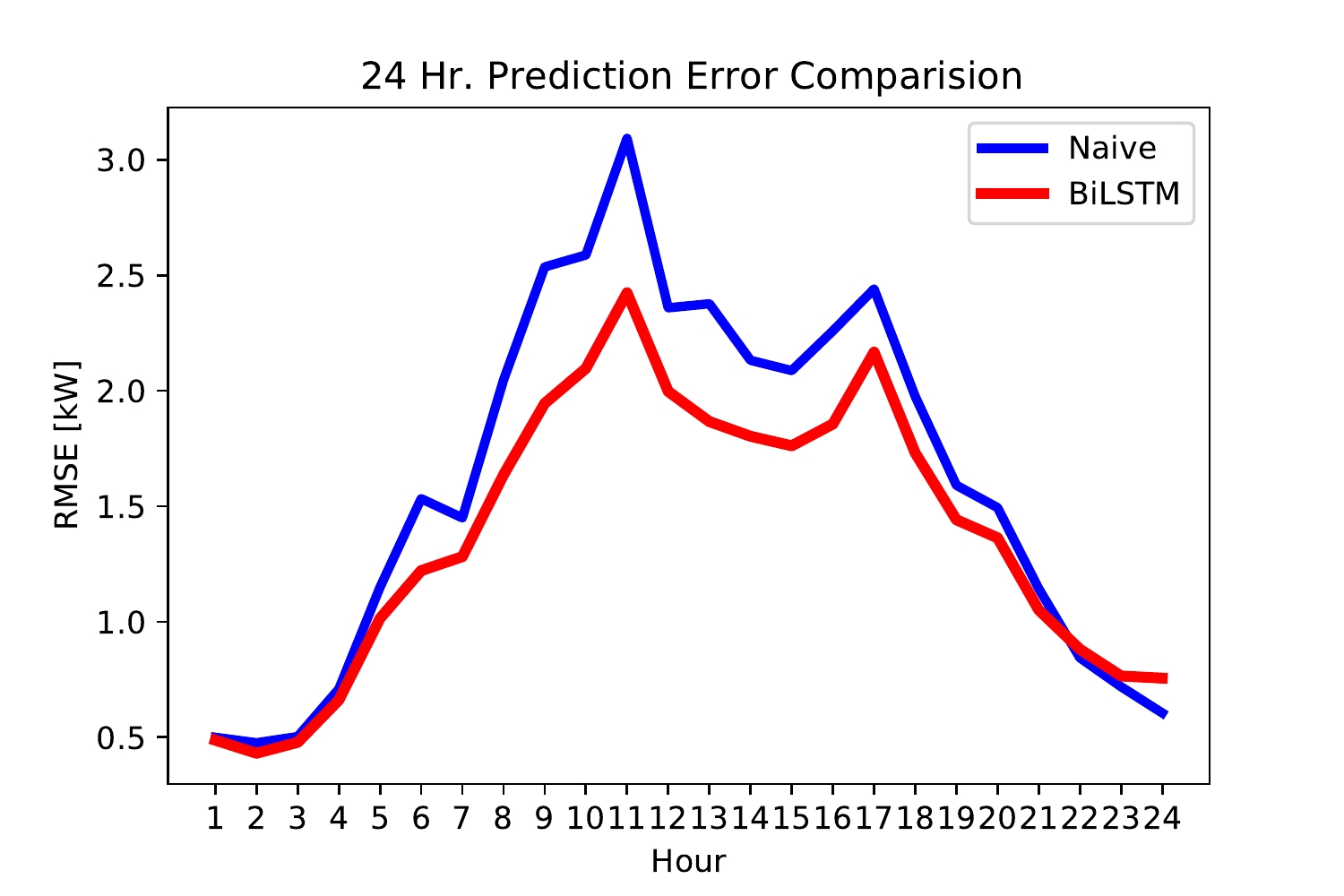}}
\subfigure[]{\includegraphics[width = 0.2412\textwidth]{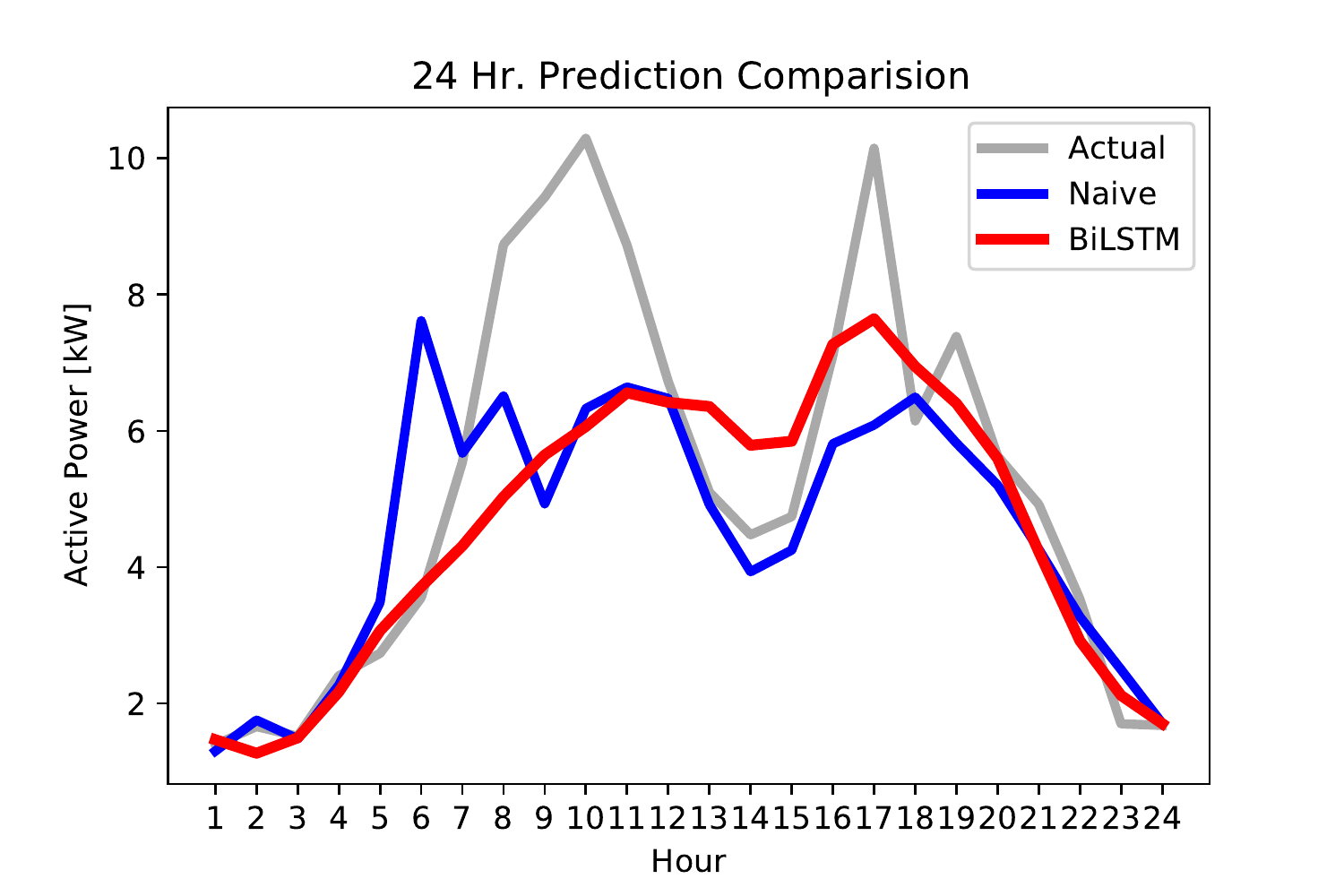}}
\captionsetup{font=small}
\caption[]{(a) Hourly RMSE, (b) Hourly load forecast  comparison of naive forecasting vs BiLSTM}
\label{fig:naive-bilstm-plots}
\end{figure}

\subsection{CNN-BiLSTM Forecast}
As shown in table \ref{tab:mystudy-comparision}, an average RMSE of the BiLSTM model is $2.6\%$ lower than that of CNN-BiLSTM. However, if additional and more complex data are involved, such as weather, renewable sources (PV, wind turbine, etc.), and individual household demand response, CNN would perform better.

\subsection{Models Comparison and Discussion}
To observe the effectiveness of the BiLSTM model, table \ref{tab:paper-comparision} compares the method to a similar study \cite{Wen2019} that uses machine learning algorithms, i.e., support vector machine (SVM), LSTM, multi-layered perceptron (MLP) to forecast daily electricity consumption at an hourly resolution for a micro-grid consisting of $38$ homes. Compared to those models, the proposed method is able to achieve an RMSE reduction of between $50.31-76.03\%$.

\begin{table}[h]
\centering
\captionsetup{font=small}
\caption{Comparison of the proposed model with the similar study \cite{Wen2019}}
\label{tab:paper-comparision}
\begin{tabular}{|c|c|c|}
\hline
Algorithms & \begin{tabular}[c]{@{}c@{}}RMSE\\ (Average)\end{tabular} & \begin{tabular}[c]{@{}c@{}}BiLSTM\\ Improvement ($\%\Delta$)\end{tabular} \\ \hline
SVM & 6.191 & 75.76 \\ \hline
MLP & 5.654 & 73.45 \\ \hline
LSTM & 2.987 & 49.75 \\ \hline
BiLSTM (this paper) & \textbf{1.501} & - \\ \hline
\end{tabular}
\end{table}
\begin{figure}[h]
\centering
\captionsetup{justification=centering}
\includegraphics[width = 0.5\textwidth]{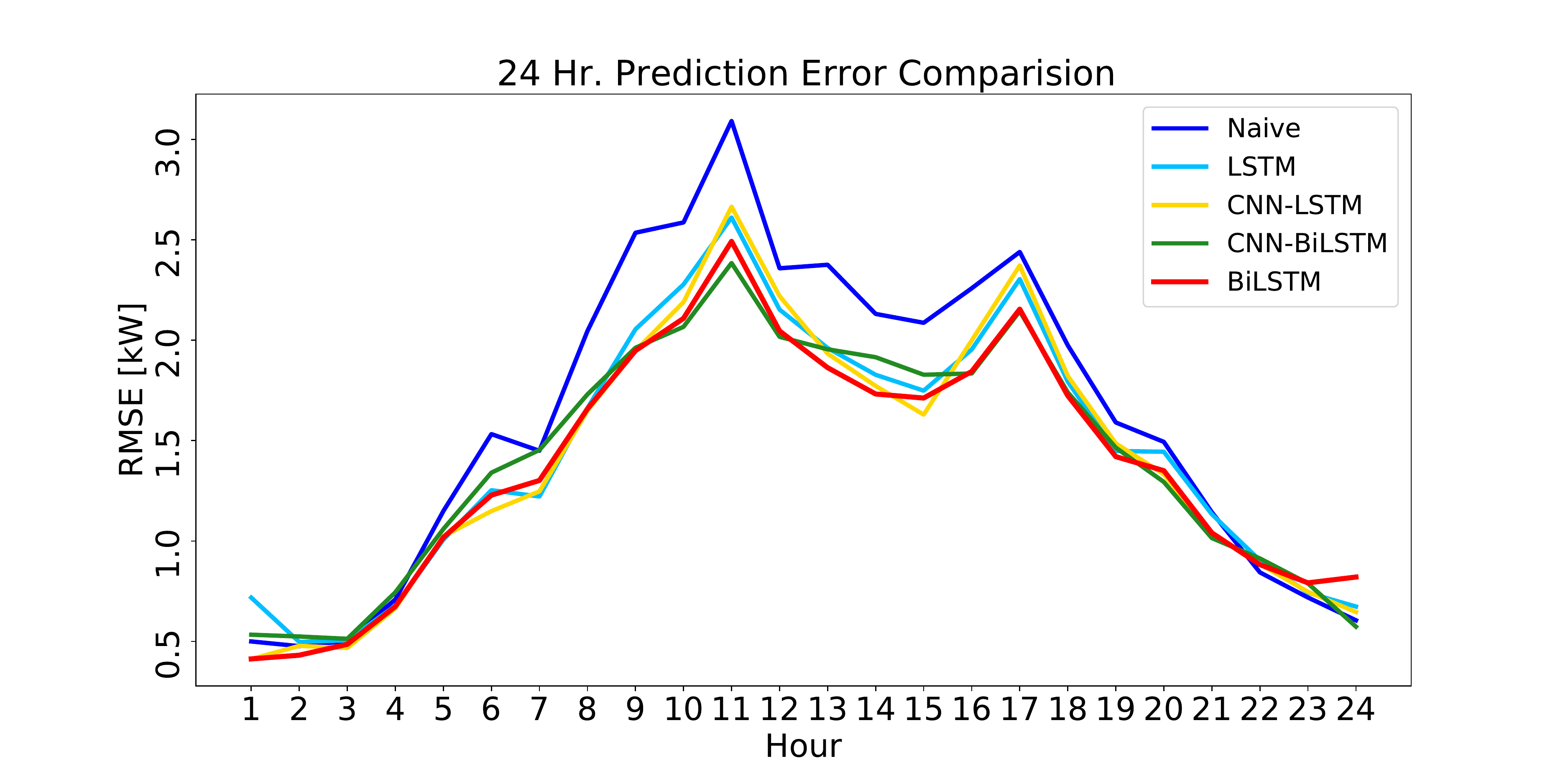}
\captionsetup{font=small}
\caption[]{Comparison of hourly RMSE of different models}
\label{fig:rmse-comparision}
\end{figure}
\begin{figure}[h]
\centering
\captionsetup{justification=centering}
\includegraphics[width = 0.5\textwidth]{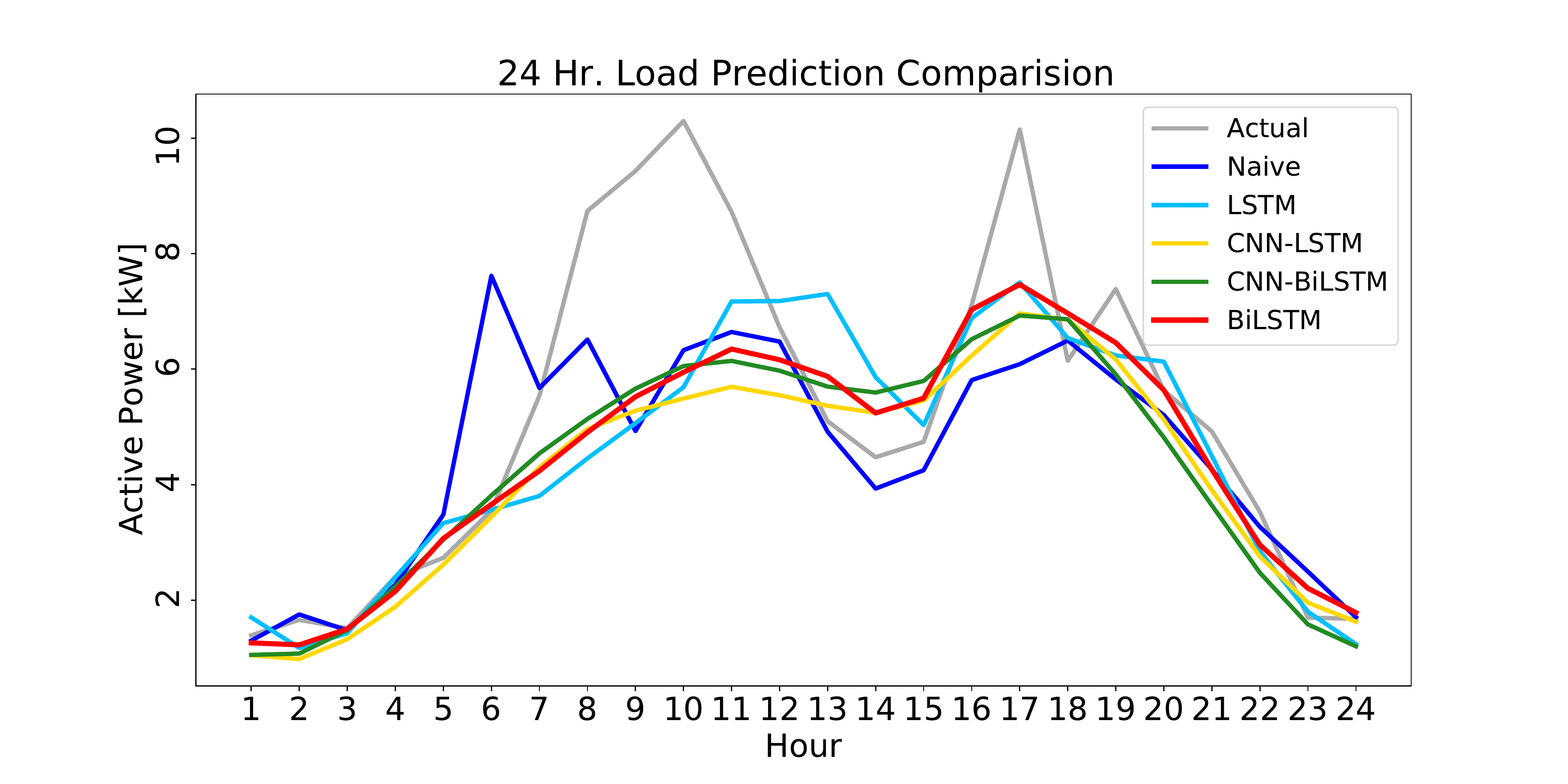}
\captionsetup{font=small}
\caption[]{Comparison of hourly prediction of different models}
\label{fig:prediction-comparision}
\end{figure}

In addition, the BiLSTM and CNN-based BiLSTM models are compared with the naive forecasting technique, improved LSTM, and CNN-LSTM models that are trained during the study. BiLSTM's RMSE is lower than any of the models, as shown in table \ref{tab:mystudy-comparision}, and fig. \ref{fig:prediction-comparision}, and observed to be $16\%$, $5.60\%$, $2.85\%$ and $2.60\%$ better than the naive forecasting, improved LSTM, CNN-LSTM and CNN-BiLSTM models respectively.

\begin{table}[t]
\centering
\captionsetup{font=small}
\caption{Comparative study of the proposed model with other frequently adopted models for dynamic load forecasting}
\label{tab:mystudy-comparision}
\begin{tabular}{|c|c|c|c|}
\hline
Algorithms & \begin{tabular}[c]{@{}c@{}}No. of trainable\\ parameters\end{tabular} & \begin{tabular}[c]{@{}c@{}}RMSE\\ (Average)\end{tabular} & \begin{tabular}[c]{@{}c@{}}BiLSTM\\ Improvement ($\%\Delta$)\end{tabular} \\ \hline
Naive & 0 & 1.787 & 16 \\ \hline
LSTM & 51,001 & 1.590 & 5.60 \\ \hline
CNN-LSTM & 332,457 & 1.545 & 2.85 \\ \hline
BiLSTM & \textbf{101,801} & \textbf{1.501} & - \\ \hline
CNN-BiLSTM & 664,457 & 1.541 & 2.60 \\ \hline 
\end{tabular}
\end{table}

The data has a strong seasonal component but lacks a trend and a substantial noise as the dataset comprises $38$ residential homes. It is possible to consider the utility of the BiLSTM network for making more accurate short-term forecasting to support the economic load dispatch of the community microgrid, similar to the paper \cite{Wen2019}. The load forecasting provided by machine learning can be used to model the load variation over a period in the future and thus, allow for matching supply and demand in community microgrid optimal load dispatch models for grid-connected community microgrids.
%
\section{Conclusion}

Different deep learning models, i.e., standard LSTM, CNN-LSTM, bidirectional LSTM (BiLSTM), and CNN-BiLSTM, are trained to perform a day ahead load forecast of the aggregated residential load of $38$ homes given the previous day load profile. The RMSE of BiLSTM architecture is observed to be $16\%$, $5.60\%$, $2.85\%$, and $2.60\%$ better than the naive forecast, LSTM, CNN-LSTM, and CNN-BiLSTM respectively. Because of the information loss during max-pooling, CNN-BiLSTM cannot perform better than the regular BiLSTM model. However, it is believed that with the increased diversity and variability in the data - more than just the seasonal pattern, the CNN-BiLSTM would perform better than BiLSTM. Hence, in the future, the performance of BiLSTM and CNN-BiLSTM models will be evaluated on more dynamic and uncertain data such as weather data and the more extensive load profiles involved.

\section{Acknowledgement}
The authors would like to thank the deHumlaTek AI Research and Mr. Saurav Adhikari for their helpful suggestions to improve the overall paper quality.

{\small
\bibliography{bilstm.bib}{}
\bibliographystyle{IEEEtran}
}
\end{document}